# DocEmul: a Toolkit to Generate Structured Historical Documents

Samuele Capobianco
Dipartimento di Ingegneria dell'Informazione
Università degli Studi di Firenze
Florence, Italy
Email: samuele.capobianco@unifi.it

Simone Marinai
Dipartimento di Ingegneria dell'Informazione
Università degli Studi di Firenze
Florence, Italy
Email: simone.marinai@unifi.it

*Abstract*—We propose a toolkit to generate structured synthetic documents emulating the actual document production process. Synthetic documents can be used to train systems to perform document analysis tasks. In our case we address the record counting task on handwritten structured collections containing a limited number of examples. Using the DocEmul toolkit we can generate a larger dataset to train a deep architecture to predict the number of records for each page.

The toolkit is able to generate synthetic collections and also perform data augmentation to create a larger trainable dataset. It includes one method to extract the page background from real pages which can be used as a substrate where records can be written on the basis of variable structures and using cursive fonts. Moreover, it is possible to extend the synthetic collection by adding random noise, page rotations, and other visual variations.

We performed some experiments on two different handwritten collections using the toolkit to generate synthetic data to train a Convolutional Neural Network able to count the number of records in the real collections.

## I. INTRODUCTION

In the last years the extensive use of Deep Learning techniques brought an improvement of results in a wide range of research areas especially in computer vision. Deep learning architectures have a large set of free parameters which have to be learned in the training phase. To achieve the best results from the training phase large datasets are an obvious need to learn these parameters. Often the labeled data are limited and in this case the training set should be extended in some way.

The generation of synthetic datasets is one common procedure when the labeled data are not large enough or when the production of ground truth is expensive. Several works use this approach to improve the results in different tasks. For instance, in [1] is proposed the creation of synthetic datasets to perform text localization in natural images. To address the ImageNet challenge, the data augmentation approach proposed in [2] consists in image translations, horizontal reflections and patch extractions.

To address document image analysis tasks the generation of synthetic document images has been adopted for many years now. The seminal work by Baird (e.g. [3]) studied the problem of document image degradation and how to model it to improve the document recognition algorithms. Using the model proposed it is possible to generate synthetic data sets used in training classifiers for document image recognition systems. These techniques have been applied particularly to address printed documents and OCR algorithms.

In the area of historical printed documents, [4] proposed one solution for word spotting by generating synthetic image words. After one suitable pre-processing step performed to segment real words on the document images, the word matching involves the comparison of one query word synthetically generated with all the indexed words.

In [5] it is proposed one relevant approach for the automatic generation of ground-truth information that can be used to design a complete system for solving document image analysis and understanding tasks. In particular, it is possible to define the document structure by means of an XML file and appropriate stylesheets. The system produces synthetic documents with ground truth informations that can be used for page segmentation, layout structure analysis and other tasks. In the area of handwriting recognition, cursive fonts have been used to synthetically generate handwritten documents [6]. More sophisticated models, not needed in our task, have been proposed as well [7].

### A. Overview

In this work we describe one open source toolkit, called *DocEmul*, that can be used to support the researchers to generate synthetic documents which emulate a real structured handwriting collection. The main focus of the system is in the generation of structured synthetic Documents i.e. documents with a record-like structure. To the best of our knowledge this is one important novelty of the proposed tool. By using the *DocEmul* toolkit it is possible to use the synthetic data to address document analysis tasks also when the real labeled datasets are too small for a suitable model training. In particular, we aim to design one generator for structured documents where each page might be composed by one header (usually at the top of the page) and several records that are written in the rest of the page. If we think of a structured document as a sort of handwritten table with different columns, every records can be considered as composed by several text lines and in turn each line can be formed by a variable number of cells (text boxes).

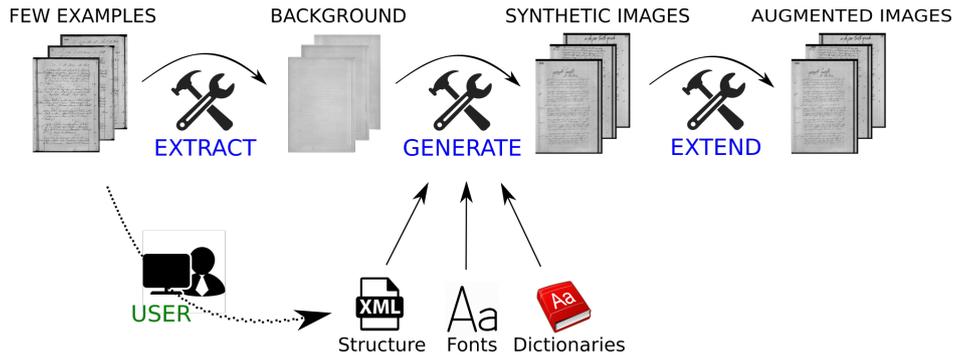

Fig. 1. Flowchart of the synthetic document production process. It is possible to set the document structure, the fonts and the dictionaries used to generate the pages.

We depict in Figure 1 one overview on all the implemented tools that can be used to generate synthetic handwritten documents. Starting from some example pages selected from the collection which we want to model, it is possible to extract the page background. This background is then used as a substrate where records will be written on the basis of variable structures using cursive fonts. Moreover, it is possible to extend the synthetic collection by adding random noise, page rotations, and other visual variations.

To indirectly evaluate the DocEmul toolkit we considered the task of record counting in historical handwritten documents by using deep architectures [8]. The target is therefore to count the number of records in each page and not to explicitly segment each record. This task can be seen a first step towards layout analysis and record segmentation.

We performed some experiments on two collections: one benchmark dataset proposed in [9] for addressing the segmentation of historical handwritten documents and one collection composed by images provided by Ancestry (the global leader in family history and consumer genomics) through our research collaboration. We propose an open source toolkit for the semi-automatic generation of synthetic handwritten documents containing records following a general structure. The Python code, together with document structures and some generated images can be freely downloaded [1].

The rest of the paper is organized as follows. In Section II we describe the toolkit developed for generating synthetic document images. In Section III we explain how to model two collections using the toolkit. Additional details on both collections will be also provided in Section III. In Section IV we summarize the experiments for the record counting task using CNNs trained with the synthetic data generated with the toolkit. Some conclusions are then drawn in Section V.

## II. A Toolkit to Generate Structured Documents

The DocEMul toolkit makes it possible to model the document aspect generating synthetic documents which look similar to actual pages. The toolkit is designed to emulate the document production process imitating the different ways adopted by the writers to create a kind of document collection.

In Figure 2 we present some examples of real documents which we want to emulate. The collections are composed by handwritten structured documents where the production process created a table-like structure composed by several columns and a variable representation of records.

In these documents we need to model different aspects of the page. Historical documents could have degradations caused by aging and storage conditions which complicate the task. To properly model these artifacts, it is useful to model also the background. In this toolkit this is achieved by extracting it directly from real images. Since several writers participated in the collection production during the time, usually we can find different writing styles and also the record structure could change with some variations from page to page. These are some of possible variations that could be seen in historical handwritten collections which make the task challenging.

The DocEmul toolkit is composed by separate modules which could work together or at different times. In the following we describe each module in details.

### A. Background Extraction

In order to extract the page background from real pages, we need to remove only the ink used to write the text on the paper and therefore no layout analysis is performed on the documents. The first step is to localize background pixels in the page by using binarization algorithms. The use of binarization in DocEmul is finalized at identifying and preserving pixels that most likely belong to the background and not at exactly identifying foreground pixels. One inaccurate binarization is therefore acceptable for the sub-sequent steps. The aim is therefore to find pixels to erase replacing them with a generated background. In the tool it is possible to use two different algorithms for the binarization task, the choice depends on the noise we want to preserve for the background extraction. The tool can use the Otsu [10] and the Sauvola [11] binarization algorithms. By identifying the foreground pixels it is possible to substitute them with suitable values that resemble the background. In particular, we "clean" the foreground pixels by replacing them with the average value of background pixels in a W × W window centered over each foreground pixels. Usually the window size is 20 × 20, but this is a changeable parameter of the tool.

[1] https://github.com/scstech85/DocEmul

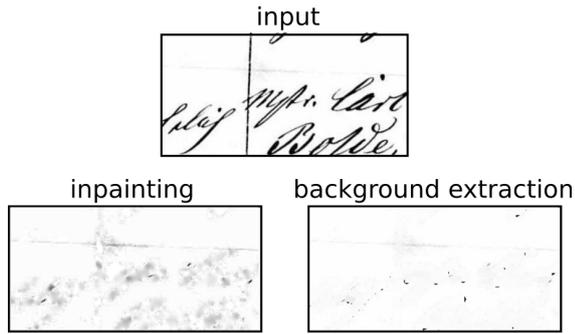

Fig. 2. An example for the background extraction step. We compare the inpainting by biharmonic functions with our approach.

By using this module, it is possible to obtain some background pages from few pages in the collection which we want to model. We have compared this approach with an image inpainting by biharmonic functions [12] from scikit-image [13], we can see in the Figure 2 as the inpainting technique produces not usable background for our task.

In Section IV we will see as the background extraction from real pages can help for training the models.

*B. Structuring Handwritten Pages*

In order to produce handwritten structured pages, the idea is to define one "ad hoc" general structure to model one collection adding random variability in the generation phase. Following the approach proposed in [14], we propose a model-driven generation technique defining a flexible model used to create synthetic documents adding some visual variations.

Having a look to real document collections which we want to emulate we are able to understand the general page structure for a given collection. Some of the features that can be inferred from the collection and that can be modeled in the tool are the fixed structure of the records and the presence of preprinted structures (e.g. vertical lines). These are examples of features which should be defined to produce synthetic pages that resemble as most as possible the real collection.

We aim to create one flexible description to define the most important document features in an easy way. We therefore use an XML file to define the configurations needed to create the synthetic pages. This file is used to characterize the header structure, the record structure, the fonts, the dictionaries, and the graphic objects which need to be used.

*1) Record and Header Structures:* We model the record as one group of text lines where each line contains several text boxes defined as cells. We can set the height for each text line and the interline spacing between consecutive text lines. It is also possible to define a probability to write the text line in the page, in this way we add some variability to the document.

Each text line contains cells which extend the text line attributes defined above. It is also possible to define other attributes for each text box. For instance, the font and the dictionary used to write the text, the horizontal position and the width of the text box. Moreover, we can add some position variability to simulate a random variation of the text flow. It is also possible to define the probability of adding one cell to the structure therefore generating records with a variable number of items according to the model.

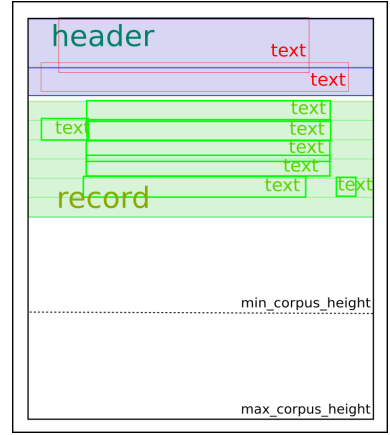

Fig. 3. Page structure defined in the XML file.

After defining the record structure it is possible to define one or more groups of records where each type of record has associated one probability of appearing in the document. In this way we can define multiple structures of records and the morphology of the pattern which we want to model. We describe the document header like other records, however, in this case it is not repeatable.

*2) Fonts and Dictionaries:* We can assign different options for each cell, in particular we can define the font to write the text in these cells. It is possible to use different fonts downloaded from web sites. We can also define a set of dictionaries to use during the production phase and we can associate one dictionary to each cell.

*3) Graphic Objects:* In order to model some types of documents it is possible to add graphic objects to the document. Using this tool we can add objects as lines or boxes that can be filled with a fixed color or with salt and pepper noise.

*C. Generating Documents*

In the document generation some areas are defined to specify the regions where we are going to print the document contents. In Figure 3 we graphically illustrate the main areas. The blue area describes the header of the page. The record of the header, like other records, can contain mandatory fields (e.g. the page number) and other fields that are added according to a specified probability distribution.

The corpus area defines the part of the page where records are printed. Any page generated by the program will contain records in the area below the header and ending between the min_corpus_height and the max_corpus_height delimiters. The area depicted in green defines the zone in the page where one record is generated.

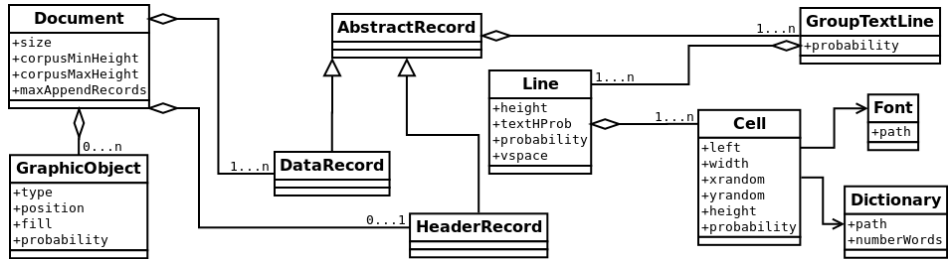

Fig. 4. UML diagram to define the document structure used by the toolkit.

## D. Data Augmentation

One module of the toolkit allows to augment an input dataset modifying real images with some artificial transformations. In particular, we can add salt and pepper noise in the whole page or in random sub areas of the page. We can randomly rotate the page choosing the angle in a fixed range $[-A°, A°]$. It is also possible to add a scale variability to the document.

## E. Model Specification

To better understand how to model one collection, we present the architecture used to define the page structure. In Figure 4 we show the UML diagram describing the XML model used to define the page structure.

We can define a document (Document) setting the size (height, width) and the corpus area between the minCorpusHeight and the maxCorpusHeight delimiters. Moreover, the tool generates a random number of records (maxAppendRecords) to fill the document in the previous defined corpus area. A document could contain graphic objects as lines or rectangles. It is possible to define them using the related class (GraphicObject) which defines the absolute position and also the probability to appear in the document.

In order to obtain an abstract document as a record container, we have defined an abstract concept for the record (AbstractRecord). This class is used to define the structure of the record. In structured documents we can have an header record (HeaderRecord) which appears at most one time for a page containing several data records (DataRecord). Using the previous definition, an abstract record is composed by several groups of text lines (GroupTextLine) which are used to define a structured record. The text lines (Line) are associated to the group, we can define the height with a random variation (textHProb) for each line and also the probability to appear during the generation.

Defining a document as a container of text lines, the tool generates a document starting from the top to the bottom of the page, for each text line we define the interline space (vspace) with respect to next text line. A text line is composed by several text boxes (Cell) to fill with text. In order to obtain more variability, a cell is associated to a random position and the probability to appear in the document. A cell is associated to a font (Font) used to write the text extracted randomly from a dictionary (Dictionary).

The generation phase creates synthetic documents composed by only text and graphic objects over a transparency layer saved in RGBA file format. It is useful to generate different versions of the same collection but with different background pages.

## III. MODELING DATASETS

We used the *Esposalles* collection (containing 200 pages) and the *Brandenburg* (containing 4,956 pages) to evaluate the toolkit capability to generate synthetic documents emulating the real collections. These are only examples of possible uses of the toolkit that is designed to be general and suitable to model also other collections.

## A. Esposalles

This collection is one benchmark dataset proposed by [9] containing historical handwritten documents. These structured documents are composed by records which we can model with our toolkit. In Figure 6 we show one example page from the dataset. For this experiments we generated 81,060 synthetic documents images containing a number of records between 3 and 9. The inputs have a resolution of size $366 \times 256$ pixels with black/white values.

In Figure 5 on the left is depicted an image for the dataset which we want to emulate. Having a look to the page, we can see how it is structured. In particular, on the top we can find the header composed by two lines. We can therefore model the header as a record with two lines where the top line has only one cell to show the page number. The second line contains the text as a title. The page contains 6 records and each record is structured in a different way. In Figure 5 we also show some extracted records and their structure. Having a look to these records we can observe that the structure for each record could change in the same page.

It is possible to simulate the production process using the toolkit and defining the document structure in the XML format and using some cursive fonts: "Scriptina", "A glitch in time" and "Love letter tw"; downloaded from http://www.dafont.com and Italian text as dictionary.

In Figure 6 we show some synthetic images generated to emulate the *Esposalles* dataset. The two images are generated with different fonts.

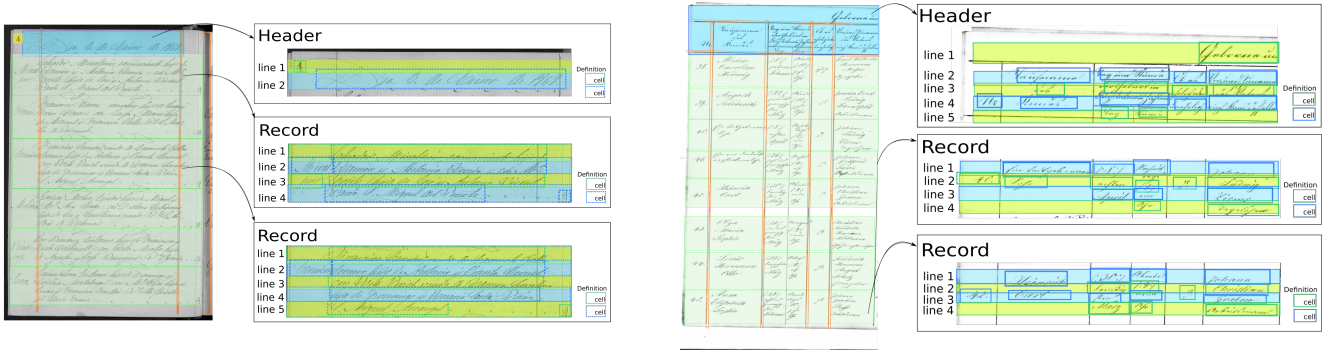

Fig. 5. Page structure. Left *Esposalles* dataset. Right *Brandenburg* dataset.

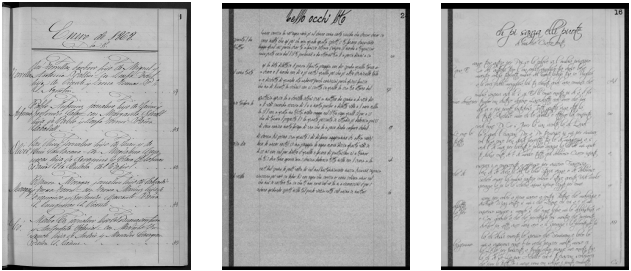

Fig. 6. *Esposalles* dataset. One real image (left) and two generated pages using the same resulution in input to the networks.

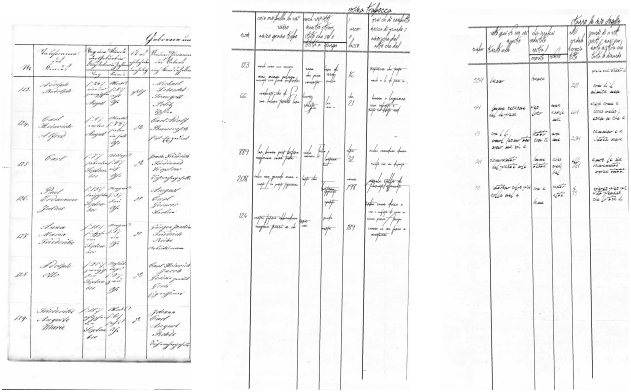

Fig. 7. *Brandenburg* dataset. One real image (left) and two generated pages

### B. Brandenburg

This collection is a private dataset from the Ancestry company and it is composed by structured documents. In Figure 7 we show one example page from the collection. For this experiments we generated $61,914$ synthetic images containing a number of records between $1$ and $10$. The input size is $450 \times 190$ pixels with black/white values.

We can observe that the document structure is more complex than in the previous collection.

In Figure 5 on the right is depicted one image from the dataset which we want to emulate. Again, on the top we can find the header record. This page contains 8 records and each record is structured differently. Moreover, we also show some extracted records and their structure.

It is possible to simulate the production process using the toolkit and defining the document structure in the XML format using various cursive fonts "Scriptina", "A glitch in time", "Love letter tw", and "Taken by vultures"; downloaded from http://www.dafont.com and Italian text as dictionary.

In Figure 7 we show some synthetic image generated to emulate the *Brandenburg* dataset.

## IV. EXPERIMENTS

To indirectly evaluate the toolkit we used it to generate training data to address the record counting in historical handwritten documents by means of deep architectures. The toolkit is used in this case to simulate the document production process using the generated data to train a Convolutional Neural Network to count the number of record for each page. We performed the experiments with two well known CNN architectures: Alexnet [2] and NetworkInNetwork [15].

To evaluate the system performance we computed two values. The Accuracy is the percentage of pages where the number of records is correctly identified. The Error is the percentage of errors in the record count when making a decision on one page at a time. The Error is defined according to Equation (1) where $r_i$ is the actual number of records in page $i$, $p_i$ is the predicted value ($\lfloor p_i + \frac{1}{2} \rfloor$ is the rounded predicted value) and $N$ is the number of test pages.

$$Error = \frac{\sum_{i=1}^{N} \left| \lfloor p_i + \frac{1}{2} \rfloor - r_i \right|}{\sum_{i=1}^{N} r_i} \quad (1)$$

### A. Esposalles

In the *Esposalles* collection we compared the results obtained by the proposed approach with those described in [9] considering the same split of the data in training, validation, and test datasets. The results are shown in Table I. As a comparison in [9] the right number of records is predicted in the $80\%$ of the test documents. It is worth mentioning that the system described in [9] is designed to segment the records

TABLE I
*Esposalles* DATASET. RESULTS ON BENCHMARK SPLIT

| Model | Input | Accuracy (%) | Error (%) |
|---|---|---|---|
| AlexNet | White BG | 85.0 | 2.6 |
| AlexNet | Extracted BG | 90.0 | 1.7 |
| NIN | Extracted BG | 95.0 | 0.9 |
| [Alvaro et al][9] | | 80.0 | - |

TABLE II
*Brandeburg* DATASET. COMPARISON OF DIFFERENT ARCHITECTURES.

| Model | Training | Accuracy (%) | Error (%) |
|---|---|---|---|
| Alexnet | Real | 66.53 | 6.19 |
| Alexnet | Synth | 46.63 | 12.40 |
| Alexnet | Synth + FT | 74.87 | 4.40 |
| NIN | Real | 74.47 | 4.53 |
| NIN | Synth | 51.26 | 9.84 |
| NIN | Synth + FT | **77.79** | **3.96** |

and therefore the record counting is only one information that is extracted from the segmentation.

We made two other experiments to evaluate the performance of the proposed method in this dataset. The first experiment is performed to evaluate the impact of the use of a white background (White BG) instead of the background extracted with the procedure described in Section II-A (Extracted BG). We compared the two approaches with the AlexNet architecture and concluded that in general it is better to use a real background than a white one. In the second experiment we compared on the same data the performance achieved with the AlexNet and NIN architectures. With these data, and also in the experiments made with the *Brandenburg* collection, the NIN slightly outperforms the AlexNet architecture.

*B. Brandenburg*

The second group of experiments is made on the 4,956 pages of the *Brandenburg* collection from Ancestry company. We split the dataset in three sets: 3,165 pages for training, 796 for validation and 995 for test. In this case we generate synthetic document to emulate the handwritten collection in order to count the number of record for each page. In this case we use the generated synthetic dataset to train CNNs and then fine tune the models with the real data. As we can see in Table II by using the pretrained model on synthetic data to fine tune the prediction model improves the results for both measures.

V. CONCLUSIONS

In this paper we have presented one open source toolkit used to generate structured handwritten documents following a record structure defined to emulate a real document collection. We have evaluated the toolkit to generate a large dataset used to train deep convolutional networks, obtaining good results in the record counting task for historical structured documents. In future work we are extending the toolkit to identify a technique to semi-automatically generating the XML configuration file starting from real documents.

ACKNOWLEDGMENTS

We would like to thank the authors of [9] for sharing the annotated dataset used to perform the experiments.
This work is partially supported by a research grant from Ancestry.